\documentclass[11pt,a4paper]{article}
\usepackage{authblk}
\usepackage[hyperref]{emnlp2020}
\usepackage{times}
\usepackage{latexsym}

\usepackage{microtype}

\usepackage{enumerate}

\usepackage{lscape}
\usepackage{afterpage}
\usepackage{capt-of}

\usepackage{booktabs}
\usepackage{tabularx}
\usepackage{multirow}
\usepackage{siunitx}
\usepackage{arydshln}
\renewcommand{\arraystretch}{1.1}

\usepackage{amsmath,amsfonts,amsthm,bm}

\usepackage{pifont}% http://ctan.org/pkg/pifont
\newcommand{\cmark}{\ding{51}}%
\newcommand{\xmark}{\ding{55}}%

\newcolumntype{Y}{>{\centering\arraybackslash}X}

\newcolumntype{R}{>{\raggedleft\arraybackslash}X}

\aclfinalcopy % Uncomment this line for the final submission
\usepackage{adjustbox}
\usepackage{array}
\usepackage{xcolor}
\title{Let's Stop Incorrect Comparisons in End-to-end Relation Extraction!}
\author[1,2]{Bruno Taill\'{e}}
\author[1]{Vincent Guigue}
\author[2]{Geoffrey Scoutheeten}
\author[1,3]{Patrick Gallinari}

\affil[1]{Sorbonne Universit\'{e}, CNRS, Laboratoire d'Informatique de Paris 6, LIP6}
\affil[2]{BNP Paribas}
\affil[3]{Criteo AI Lab}
\affil[ ]{\texttt{\{bruno.taille, vincent.guigue, patrick.gallinari\}@lip6.fr}}
\affil[ ]{\texttt{geoffrey.scoutheeten@bnpparibas.com}}

\usepackage{fancyhdr}
\usepackage{geometry}
\begin{document}
\maketitle
\begin{abstract}
Despite efforts to distinguish three different evaluation setups \cite{bekoulis-etal-2018-adversarial, Bekoulis2018JointProblem}, numerous end-to-end Relation Extraction (RE) articles present unreliable performance comparison to previous work.
In this paper, we first identify several patterns of invalid comparisons in published papers and describe them to avoid their propagation.
We then propose a small empirical study to quantify the most common mistake's impact and evaluate it leads to overestimating the final RE performance by around 5\% on ACE05.
We also seize this opportunity to study the unexplored ablations of two recent developments: the use of language model pretraining (specifically BERT) and span-level NER.
This meta-analysis emphasizes the need for rigor in the report of both the evaluation setting and the dataset statistics.
We finally call for unifying the evaluation setting in end-to-end RE \footnotemark[1].
\end{abstract}

\section{Introduction}
Named Entity Recognition (NER)\footnotemark[2] and Relation Extraction (RE) are 
key Information Extraction tasks, for example at the heart of Knowledge Graph Construction along with Coreference Resolution and Entity Linking.
In the traditional pipeline approach, these tasks are treated with two 
models trained separately and applied sequentially \cite{Bach2007AExtraction}.
Nevertheless, combining information from both submodules is beneficial \cite{roth-yih-2002-probabilistic} and end-to-end RE models tackling both tasks jointly have been proposed to better model their interdependency and overcome
cascading errors \cite{li-ji-2014-incremental}.

This end-to-end setting has recently received more attention in the wake of improved language models (LM).
However, in this prolific and competitive domain, authors have used several evaluation settings to compare their performance.
And despite the attempt to clearly identify three main setups \cite{bekoulis-etal-2018-adversarial, Bekoulis2018JointProblem}, 
this multiplication of settings makes the apprehension of the literature difficult and confusing, but more importantly, it has led to erroneous comparisons and conclusions.

In this paper, we first present a quick literature review of the recent advances in end-to-end RE.
Our main contribution is then the identification of invalid comparison patterns in recent publications.
We list them with the hope of stopping the propagation of erroneous results and presenting a curated list of published results.
To further this contribution, we propose a small empirical study to quantify the impact of switching the two main metrics and estimate it can lead to a relative overestimation of around 5\% in the end-to-end RE results on ACE05.

As a second contribution, we take advantage of this quantitative study to perform the omitted ablations of two recent developments in the literature: LM pretraining and Span-level NER.
It confirms that recent empirical gains are mainly due to LM pretraining, while there is no evidence for quantitative gains from Span-level NER 
%%V2
on non-overlapping entities.

Finally, we argue that the main cause for previously identified mistakes is the lack of reproducibility and, consequently, of previous work reproductions.
We call for a more rigorous report of both evaluation settings and dataset statistics in general and particularly in end-to-end RE.
And we also suggest unifying our evaluation setting to reduce the chance of future mistakes and enable more meaningful cross-dataset analyses.

\footnotetext[1]{Code available at \href{https://github.com/btaille/sincere}{github.com/btaille/sincere}}
\footnotetext[2]{We will also use NER to refer to Entity Mention Detection (EMD) when entities of interest are not Named Entities.}

\renewcommand{\tabcolsep}{3pt}

\begin{table*}[h]
\centering
\small
% \scriptsize
\begin{tabularx}{\textwidth}{@{}l*{3}{c}l*{6}{c}lccll@{}}

\toprule
    &  & &       &       & \multicolumn{6}{c}{Representations} & & Enc. & \multicolumn{2}{c}{NER} & RE \\
            \cline{6-11} \cline{14-15}

% Reference   &  Crit. & Code & LM & Word & Char & Hand & POS & DEP  \\
Reference   &  \multicolumn{2}{c}{Criterion} & Code & & LM & Word & Char & Hand & POS & DEP & & &  Tag & Dec. & Dec.\\
\midrule

\cite{Giorgi2019End-to-endModels} & S & & \href{https://github.com/bowang-lab/joint-ner-and-re}{*}&
                                            & B & & & & & &
                                            & - 
                                            & B & MLP
                                            & Biaff. \\
                                            
\cite{Eberts2020Span-basedPre-training} & SB &\cmark & \href{https://github.com/markus-eberts/spert}{\cmark} &
                                            & B & & & & & &
                                            & - 
                                            & S& MLP
                                            & PMaxPool\\
                                            
\cite{Wadden2019EntityRepresentations} & B &\cmark &\href{https://github.com/dwadden/dygiepp}{\cmark} &
                                            & B & & & & & &
                                            & - 
                                            & S& MLP
                                            & Biaff. \\ 
                                            
\cite{li-etal-2019-entity} & S & &   &
                                            & B & & & & &  &
                                            & -  
                                            & - & MT QA & MT QA \\
                                            
\cite{Dixit2019Span-LevelExtraction} & S & &  &
                                            & E & S & C & & & &
                                            & L 
                                            & S& MLP
                                            & Biaff.\\
                                            
\cite{luan-etal-2019-general} & B& \cmark & \href{https://github.com/luanyi/DyGIE}{\cmark}  &
                                            & E & G & ns & & & &
                                            & L 
                                            & S & MLP
                                            & Biaff.\\
                                            
\cite{Nguyen2019End-to-endAttention} & SR& \cmark & \href{https://github.com/datquocnguyen/jointRE}{\cmark} &
                                            & & G & L & & &  &
                                            & L 
                                            & B& MLP
                                            & MHS-Biaff.\\
                                            
\cite{Sanh2019ATasks} & - &\xmark & \href{https://github.com/huggingface/hmtl}{\cmark}&
                                            & E & G & C & & &  &
                                            & L 
                                            & B & CRF 
                                            & MHS-Lin.\\
                                            
\cite{luan-etal-2018-multi} &  B &  & \href{http://nlp.cs.washington.edu/sciIE/}{\cmark} &
                                            & E & G & ns & & &  &
                                            &   
                                            & S & MLP
                                            & Biaff.\\
                                            
\cite{sun-etal-2018-extracting} & S & \cmark & \href{https://github.com/changzhisun/entrel-joint-mrt}{$\approx$} &
                                            & & ns & C & & &  &
                                            & L  
                                            & B & MLP
                                            & PCNN\\
                                            
\cite{bekoulis-etal-2018-adversarial, Bekoulis2018JointProblem} & SBR & \cmark & \href{https://github.com/bekou/multihead_joint_entity_relation_extraction}{\cmark}&
                                            & & S/W & L & & &  &
                                            & L 
                                            & B & CRF 
                                            & MHS-Lin.\\

\cite{zhang-etal-2017-end} & S & \cmark & \href{https://github.com/zhangmeishan/NNRelationExtraction}{$\approx$}&
                                            & & G & C & & \cmark & \cmark &
                                            & L 
                                            & B & I-LSTM & I-LSTM \\

\cite{Li2017AText} & S & \cmark & \href{https://github.com/foxlf823/njmere}{$\approx$} &
                                            & & ns & C & & \cmark & \cmark  &
                                            & L 
                                            & B & MLP
                                            & SP LTSM\\

\cite{katiyar-cardie-2017-going} & S & \cmark &  &
                                            & & W & & & &  &
                                            & L 
                                            & B & I-MLP & I-Pointer\\

\cite{Zheng2017JointNetwork} & S & \cmark &  &
                                            & & ns & & & &  &
                                            & L 
                                            & B & MLP
                                            & PCNN\\

\cite{adel-schutze-2017-global} & R & \cmark & \href{https://github.com/heikeadel/global_normalization}{\cmark}&
                                            & & W & & & & &
                                            & - 
                                            & B & CNN 
                                            & PCNN+CRF \\

\cite{gupta-etal-2016-table} & R & \cmark &   &
                                            & & T & & \cmark &  \cmark & &
                                            & - 
                                            & B & I-RNN & I-RNN \\

\cite{miwa-bansal-2016-end} & S & \cmark & \href{https://github.com/tticoin/LSTM-ER}{\cmark} &
                                            & & ns & & & \cmark & \cmark &
                                            & L 
                                            & B & MLP
                                            & SP LSTM\\

\cite{miwa-sasaki-2014-modeling} & S & \cmark & \href{https://github.com/tticoin/JointER}{\cmark} &
                                            & & & & \cmark & & &
                                            & - 
                                            & B & I-SVM & I-SVM \\

\cite{li-ji-2014-incremental} & SB & \cmark &  &
                                            & & & &  \cmark & & &
                                            & - 
                                            & B & I-Perc. & I-Perc. \\

\bottomrule

\end{tabularx}

\caption{Proposed classification of end-to-end RE models in antichronological order.\\
Criterion: \textbf{S}trict / \textbf{B}oundaries / \textbf{R}elaxed and presence of statement (\xmark : incorrectly stated). 
Code: source code availability (${\approx}$ : no documentation / *:WIP).
Language Model pretraining:  \textbf{E}LMo \cite{peters-etal-2018-deep} / \textbf{B}ERT \cite{Devlin-etal-2019-bert}.
Word embeddings: \textbf{S}ENNA \cite{Collobert2011NaturalScratch} / \textbf{W}ord2Vec \cite{Mikolov2013DistributedCompositionality} / \textbf{G}loVe \cite{pennington-etal-2014-glove} / \textbf{T}urian \cite{turian-etal-2010-word}.
Character embeddings pooling: \textbf{C}NN / (Bi)\textbf{L}STM.
Hand: handcrafted features. POS/DEP: use of Ground Truth or external Part-of-Speech tagger or Dependency Parser.
Encoder: (Bi)\textbf{L}STM. NER Tag: \textbf{B}ILOU / \textbf{S}pan. Decoders: I- = Incremental, MHS=Multi-Head Selection, SP=Shortest Dependency Path. ns=Not Specified, for words it might be randomly initialized embeddings.
}

\label{table:cls}
\end{table*}

\section{A Quick Literature Review}

In order to have a global view of recent evolutions, we present a quick literature review of end-to-end RE models. 
We focus on supervised extraction of intra-sentence binary relations in English corpora. A summary is proposed in Table \ref{table:cls}.

\paragraph{Local classifiers}
The first attempts to model the interdependency between NER and RE combined the predictions of independent local classifiers according to global constraints 
(e.g. the arguments of the ``Live In" relation must be a Person and a Location); 
either with Probabilistic Graphical Models \cite{roth-yih-2002-probabilistic}, Integer Linear Programming \cite{roth-yih-2004-linear} or Card Pyramid Parsing \cite{kate-mooney-2010-joint}.

\paragraph{Incremental Joint Training}
\citet{li-ji-2014-incremental} propose the first joint model using a structured perceptron to parse a sentence with a set of
% a sequence of
two actions: append a mention to detected entities and possibly link it with a relation to a previous mention.
\citet{katiyar-cardie-2017-going} adopt the same framing but replace handcrafted features with word embeddings and use a BiLSTM for NER and a Pointer Network for RE.
\citet{miwa-sasaki-2014-modeling} simplify this setting by sequentially filling a table containing all entity and relation information.
\citet{gupta-etal-2016-table} take up this Table Filling (TF) approach but use an RNN with a multitask approach.
Similarly, \cite{zhang-etal-2017-end} use LSTMs but add syntactic features from \cite{Dozat2017DeepParsing}'s Dependency Parser.
% These last two models incorporate elements of the Piecewise models introduced below.

\paragraph{Entity Filtering}
Other models use entity filtering as in the pipeline setting where RE is viewed as classification given a sentence and a pair of arguments.
This requires passing each pair of candidate entities through the RE classifier.
The only difference is that the NER and RE models share some parameters in end-to-end RE, often in a BiLSTM encoder.
Indeed, as in the previous incremental setting, NER is modeled as sequence labeling using BILOU tags \cite{ratinov-roth-2009-design} and the NER module is often a BiLSTM as in \cite{Huang2015BidirectionalTagging}.
The two modules are jointly trained by optimizing for the (weighted) sum of their losses.

\citet{miwa-bansal-2016-end} use a sequential BiLSTM for NER and a Tree-LSTM over the shortest dependency path between candidate arguments given by an external parser and
\citet{Li2017AText} apply this model to biomedical data.

\citet{adel-schutze-2017-global}, \citet{Zheng2017JointNetwork} and \citet{sun-etal-2018-extracting}  all rely on the Piecewise CNN (PCNN) architecture for RE \cite{zeng-etal-2015-distant}.
The sentence is split into three pieces: before the first argument, between the arguments, and after the last argument.
The RE classifier is fed with CNN pooled representations of these three pieces and of both arguments.
% \cite{adel-schutze-2017-global}, \cite{zheng-etal-2017-joint} and \cite{sun-etal-2018-extracting}  all use a PCNN.
\citet{adel-schutze-2017-global} add a CRF to model the argument type / relation type dependencies while \citet{sun-etal-2018-extracting} use minimum risk training to incorporate global F1 scores in the loss and make loss functions more interdependent.

\paragraph{Multi-Head Selection}
To avoid relying explicitly on NER prediction, \citet{Bekoulis2018JointProblem, bekoulis-etal-2018-adversarial}, propose Multi-Head Selection where RE classification is made for every pair of words.
%VINCETN TF +> table filling (la definition était trop loin)
As in Table Filling, relations should only be predicted between the last words of entity mentions to avoid redundancy and inconsistencies.
This enables end-to-end RE in a single pass, but contextual information must be implicitly encoded in all word representations since the Linear RE classifier is only fed with representations of both arguments and a label embedding of BILOU NER predictions.
% While this enables to make the complete prediction in a single pass, contextual information must be implicitly encoded in all word representations since the Linear RE classifier is only fed with representations of both arguments and a label embedding of NER predictions.
\citet{Nguyen2019End-to-endAttention} replace this linear RE classifier by the bilinear scorer from \citet{Dozat2017DeepParsing}'s Dependency Parser.
A similar architecture is extended with BERT representations in \cite{Giorgi2019End-to-endModels}.
Finally, \citet{Sanh2019ATasks} build on \cite{Bekoulis2018JointProblem} to explore a broader multitask setting incorporating Coreference Resolution (CR) and another corpus for NER. 
They use ELMo contextualized embeddings \cite{peters-etal-2018-deep}.

\paragraph{Span-level NER}
With the same idea of jointly training CR along with joint NER and RE, \citet{luan-etal-2018-multi} replace the traditional sequence labeling framing of NER by span-level classification inspired by end-to-end CR \cite{lee-etal-2017-end} and Semantic Role Labeling (SRL) \cite{he-etal-2018-jointly}. 
In this setting, all spans (up to a fixed length) are independently classified as entities, which enables detecting overlapping entities,
and they use an element-wise biaffine RE classifier to classify all pairs of detected spans. 
In \cite{luan-etal-2019-general}, they then propose to iteratively refine predictions with dynamic graph propagation of RE and CR confidence scores. 
This work is adapted with BERT as an encoder in \cite{Wadden2019EntityRepresentations}.

\citet{Dixit2019Span-LevelExtraction} use a model very similar to \citet{luan-etal-2018-multi}'s but restrict to end-to-end RE.
\citet{Eberts2020Span-basedPre-training} recently use span-level NER with BERT as an encoder.
They add a pooled representation of the middle context for RE, similarly to piecewise models.

\paragraph{Question Answering}
RE can also be framed as Question Answering (QA) in the zero-shot \cite{levy-etal-2017-zero} or end-to-end \cite{li-etal-2019-entity} settings.
The latter Multi-Turn QA uses templates of questions to identify entity mentions and their relations.

\section{Datasets and Metrics}
\paragraph{Datasets}

Although a variety of datasets have been used, we limit our report to the five we identified as the most frequently studied for brevity.

Following \cite{roth-yih-2002-probabilistic}, end-to-end RE has traditionally been explored on English news articles, which is reflected in the domain of its historical benchmarks, CoNLL04 and the ACE datasets. 
\textbf{CoNLL04} \cite{roth-yih-2004-linear} is annotated for four entity types and five relation types and specifically only contains sentences with at least one relation. 
The \textbf{ACE04} dataset \cite{doddington-etal-2004-automatic} defines seven coarse 
entity types and seven relation types.
\textbf{ACE05} resumes this setting but merges two relation types leading to six of them.

More recently, \citet{Gurulingappa2012DevelopmentReports} propose the \textbf{ADE} dataset in the biomedical domain, which focuses on one relation, the Adverse Drug Event between a Drug and one of its Adverse Effects.
In the scientific domain, \citet{luan-etal-2018-multi} introduce \textbf{SciERC} composed of 500 scientific article abstracts annotated with six types of scientific entities, coreference clusters, and seven relations between them.

\paragraph{Metrics}
The traditional metrics for assessing both NER and RE performance are Precision, Recall and F1 scores.
However, there are two points of attention: the use of micro or Macro averaged metrics across types and the criterion used to consider a prediction as true positive.

On this second point, there is no difficulty for NER where the consensus is to both consider detection and typing.
However, compared to the pipeline Relation Classification, this end-to-end RE setting adds a source of mistake in the identification of arguments.
And while there is an agreement that the relation type must be correctly detected, several evaluation settings have been introduced with different argument detection requirements.

Hence, \citet{bekoulis-etal-2018-adversarial} distinguishes three evaluation settings: 

\textit{\textbf{Strict}}: both the boundaries and the entity type of each argument must be correct.

\textit{\textbf{Boundaries}}: argument type is not considered and boundaries must be correct.

\textit{\textbf{Relaxed}}: NER is reduced to Entity Classification i.e. predicting a type for each token. A multi-token entity is considered correct if at least one token is correctly typed.

    \begin{table*}[h]
    \centering
    \small

    \begin{tabularx}{\textwidth}{@{}l*{2}{Y}r*{2}{Y}r*{2}{Y}r*{2}{Y}r*{2}{Y}c@{}}
    
    \toprule
    & \multicolumn{2}{c}{ACE 05} &  & \multicolumn{2}{c}{ACE 04} & & \multicolumn{2}{c}{CoNLL04} & &\multicolumn{2}{c}{ADE} & &\multicolumn{2}{c}{SciERC} &\\
     \cmidrule{2-3} \cmidrule{5-6} \cmidrule{8-9} \cmidrule{11-12} \cmidrule{14-15}
    Reference  & Ent & Rel & & Ent & Rel & & Ent & Rel & & Ent & Rel & & Ent & Rel\\

    \midrule
    \midrule
    
    \textbf{Strict Evaluation}
    & \multicolumn{2}{c}{$\mu$F1} &
    & \multicolumn{2}{c}{$\mu$F1} &
    & \multicolumn{2}{c}{$\mu$F1} &
    & \multicolumn{2}{c}{$\mu$F1} & %\multicolumn{2}{c}{MF1} &
    & \multicolumn{2}{c}{$\mu$F1} \\

    % BERT EMD + Biaffine Parser  
    \cite{Giorgi2019End-to-endModels}
    % & BERT          
    & \textbf{87.2}$^{\dagger}$ & 58.6$^{\dagger}$ &   
    & \textbf{87.6}$^{\dagger}$ & \textbf{54.0}$^{\dagger}$ &
    % & 89.5$^{\ast}$ & 66.8$^{\ast}$ &
    & 89.5$^{\dagger}$ & 66.8$^{\dagger}$ &
    & \textbf{89.6} & \textbf{85.8} & 
    & & &
    \textbf{+} \\

    % SpERT               
    \cite{Eberts2020Span-basedPre-training}                 
    % & BERT          
    & - & - &   
    & - & - &
    & \textbf{88.9}$^{\dagger}$ & \textbf{71.5}$^{\dagger}$ &
    & 88.9$^{\dagger}$ & 79.2$^{\dagger}$& % 89.3$^{\dagger}$ & 79.2$^{\dagger}$&
    & & &
    -\\
                        
    % Span-level Model   
    \cite{Dixit2019Span-LevelExtraction}                    
    % & ELMo + SENNA + charCNN
    & 86.0 & \textbf{62.8} &   
    & - & - &
    & - & - &
    & - & - &
    & & &
    -\\
                                                            
    % Multi Turn QA       
    \cite{li-etal-2019-entity}
    % & BERT
    & 84.8 & 60.2 &
    & 83.6 & 49.4 &
    & 87.8$^{\ast}$ & 68.9$^{\ast}$ & 
    & - & - &
    &  &  &
    \textbf{+}\\
    
    \cdashline{1-16}
    % \cdashline{1-12}
     
    % Minimum Risk Training (PCNN)
    \cite{sun-etal-2018-extracting}                    
    % & GloVe + charCNN
    & 83.6 & 59.6 &
    & - & - &
    & - & - &
    & - & - &
    & & &
    \textbf{+}\\

    % Multi Head Selection+AT$^{\dagger}$   
    \cite{bekoulis-etal-2018-adversarial}
    % & GloVe / W2V + charBiLSTM
    & - & - &
    & 81.6$^{\dagger}$ & 47.5$^{\dagger}$ &
    & 83.6$^{\dagger}$ & 62.0$^{\dagger}$ &
    & 86.7 & 75.5 &
    & & &
    \textbf{+}\\
                            
    % Multi Head Selection$^{\dagger}$    
    \cite{Bekoulis2018JointProblem}
    % & GloVe / W2V + charBiLSTM
    & - & - &
    & 81.2$^{\dagger}$ & 47.1$^{\dagger}$ & 
    & 83.9$^{\dagger}$ & 62.0$^{\dagger}$ &
    & 86.4 & 74.6 &
    & & &
    \textbf{+}\\
    
    % Global Optimization (DEP Enc + PLSTM) 
    \cite{zhang-etal-2017-end}
    % & GloVe + charBiLSTM + POS
    & 83.6 & 57.5 &
    & - & - &
    & 85.6$^{\ast}$ & 67.8$^{\ast}$ & 
    & - & - &
    & & &
    -\\
    
     % Multi Head Selection$^{\dagger}$    
    \cite{Li2017AText}
    % & GloVe / W2V + charBiLSTM
    & - & - &
    & - & - &
    & - & - &
    & 84.6 & 71.4 &
    & & &
    \textbf{+}\\
                            
    % BiLSTM + Pointer Net    
    \cite{katiyar-cardie-2017-going}
    % & W2V
    & 82.6 & 53.6 &
    & 79.6 & 45.7 &
    & - & - &
    & - & - &
    & & &
    -\\
                            
     % Multi Head Selection$^{\dagger}$    
    \cite{Li2016JointText}
    % & GloVe / W2V + charBiLSTM
    & - & - &
    & - & - &
    & - & - &
    &  79.5 & 63.4 &
    & & &
    -\\
                            
    % CoType                  
    % & \cite{Ren2016CoType:Bases} 
    % & 
    % & - & - &  
    % & - & - &
    % & - & - & 
    % & 66.0 & 46.3 \\
                            
    % Tree LSTM               
    \cite{miwa-bansal-2016-end}
    % & W2V + POS + DEP
    & 83.4 & 55.6 &
    & 81.8 & 48.4 &
    & - & - &
    & - & - &
    & & &
    -\\
                            
    % Table Filling           
    \cite{miwa-sasaki-2014-modeling}
    % & Handcrafted local and global features
    & - & - &
    & - & - &
    & 80.7$^{\ast}$ & 61.0$^{\ast}$ &
    & - & - &
    & & &
    -\\
    
    \cite{li-ji-2014-incremental}
    % & Handcrafted local and global features
    & 80.8 & 49.5 &
    & 79.7 & 45.3 &
    & - & - &
    & - & - &
    &  &  &
    -\\
    
    \midrule
    \textbf{Boundaries Evaluation} \\
    
    % SpERT               
    \cite{Eberts2020Span-basedPre-training}                 
    % & BERT          
    & - & - &   
    & - & - &
    &  &  &
    &  &  &
    & \textbf{70.3}$^{\dagger}$ & \textbf{50.8}$^{\dagger}$&
    -\\
    
    % DyGIE ++                    
    \cite{Wadden2019EntityRepresentations} \xmark
    % & BERT          
    & \textbf{88.6} & \textbf{63.4} &   
    & - & - &
    &  &  &
    &  &  &
    & 67.5 & 48.4 &
    \textbf{+}\\
    
    \cite{luan-etal-2019-general} \xmark
    % & BERT          
    & 88.4 & 63.2 &   
    & \textbf{87.4} & \textbf{59.7} &
    &  &  &
    &  &  &
    & 65.2 & 41.6 &
    \textbf{+}\\

    % \cdashline{1-12}
    
    \cite{luan-etal-2018-multi}
    % & GloVe 6B 
    & - & - &
    & - & - &
    &  &  &
    &  &  &
    & 64.2 & 39.3 &
    \textbf{+}\\
    
    \cdashline{1-16}
    
    % Hybrid Network         
    \cite{Zheng2017JointNetwork} \xmark
    % & Handcrafted local and global features
    & - & 52.1 &
    & - & - &
    &  &  &
    &  &  &
    & - & - &
    -\\
    
    % Incremental           
    \cite{li-ji-2014-incremental}
    % & Handcrafted local and global features
    & 80.8 & 52.1 &
    & 79.7 & 48.3 &
    &  &  &
    &  &  &
    & - & - &
    -\\
    
    \midrule
    \midrule
    
    \textbf{Relaxed Evaluation}
    &  &  &
    &  &  &
    & \multicolumn{2}{c}{MF1} \\
    
    % Biaffine Parser \footnotemark[1]    
    \cite{Nguyen2019End-to-endAttention} \xmark
    % & GloVe + charBiLSTM
    &  &  &
    &  &  & 
    & \textbf{93.8} & \textbf{69.6} &
    &  & &
    & & &
    -\\
    
    % Multi Head Selection+AT$^{\dagger}$   
    \cite{bekoulis-etal-2018-adversarial}
    % & GloVe / W2V + charBiLSTM
    &  &  &
    &  &  &
    & 93.0$^{\dagger}$ & 68.0$^{\dagger}$ &
    & & &
    & & &
    \textbf{+}\\
    
    % Multi Head Selection$^{\dagger}$   
    \cite{Bekoulis2018JointProblem}
    % & GloVe / W2V + charBiLSTM
    &  &  &
    &  &  &
    & 93.3$^{\dagger}$ & 67.0$^{\dagger}$ &
    & & & 
    & & & 
   \textbf{+}\\
                            
    % Global Normalization of CNN     
    \cite{adel-schutze-2017-global}
    % & GloVe
    &  &  &
    &  &  &
    & 82.1 & 62.5 &
    &  & &
    & & &
    -\\
                            
    % Table Filling MT RNN$^{\dagger}$    
    \cite{gupta-etal-2016-table}
    % & 7-gram 50d (Turian 2010) + POS + hand
    &  &  & 
    &  &  & 
    & 92.4$^{\dagger}$ & \textbf{69.9}$^{\dagger}$ &
    & & &
    & & &
    -\\
    
    \midrule
    \midrule
    \textbf{Not Comparable}\\
    
    % Hierarchical Multi Task Learning \footnotemark[2]       
    \cite{Sanh2019ATasks} \xmark
    % & ELMo + GloVe + charCNN 
    & 85.5 & 60.5 &
    &  &  &
    &  &  &
    &  &  &
    &  &  &
    -\\
    \cdashline{1-16}                              
    \bottomrule
    
    \end{tabularx}
    \caption{Summary of recently published results in end-to-end RE on five datasets.\\
    $^{\ast}$ = partition from \cite{miwa-sasaki-2014-modeling}. $^{\dagger}$ = explicit use of train+dev. \textbf{+} = experiments on additional datasets.\\
    \xmark = some results were incorrectly reported as Strict. Models over the dashed lines use LM pretraining.}
    \label{table:re_sota}
    \end{table*}

\section{Identified Issues in Published Results}
This variety of evaluation settings, visible in Table~\ref{table:cls}, leads to confusion which in turn favors recurring mistakes.
By a careful examination of previous work and often only thanks to released source codes and/or sufficiently detailed descriptions, we identified several of them.
Because these precious sources of information are sometimes missing, we cannot assert we are exhaustive. 
However, we will now list them to avoid their propagation and present a curated summary of supposedly comparable results in Table~\ref{table:re_sota}.

\subsection{Comparing Boundaries to Strict results on ACE datasets}
The most common mistake is the comparison of Strict and Boundaries results. 
Indeed, several works  \cite{Zheng2017JointNetwork, luan-etal-2019-general, Wadden2019EntityRepresentations} use the Boundaries setting to compare to previous Strict results. However, because the Strict setting is more restrictive, this leads to overestimating the benefit of the proposed model over previous SOTA.
We propose a quantification of the resulting improper gain in section \ref{sec:boundvsstrict}.

\subsection{Confusing Settings on CoNLL04}
On the CoNLL04 dataset, the two settings that have been used are even more different.
Indeed, while \citet{miwa-sasaki-2014-modeling} use the Strict evaluation, \citet{gupta-etal-2016-table}, who build upon the same Table Filling idea, introduce a different setting.
They 1) use the Relaxed criterion; 2) discard the ``Other" entity type; 3) release another train / test split; 4) use Macro-F1 scores.

This inevitably leads to confusions, first on the train / test splits, e.g. \citet{Giorgi2019End-to-endModels} 
% , li-etal-2019-entity} 
%stating that they 
claim to use the splits from \cite{miwa-sasaki-2014-modeling} while they link to \cite{gupta-etal-2016-table}'s. 
Second, \citet{Nguyen2019End-to-endAttention} unconsciously introduce a different \textit{Strict setup} because it ignores the ``Other" entity type and considers Macro-F1 instead of micro-F1 scores. This leads to unfair comparisons.

\subsection{Altering both Metrics and Data}
\citet{Sanh2019ATasks} propose a multitask Framework for NER, RE and CR and use ACE05 to evaluate end-to-end RE.
However, they combine two mistakes: incorrect metric comparison and dataset alteration.
First, they use the typical formulation to describe a Strict setting but, in fact, use a setting looser than Boundaries.
Indeed, they do not consider the type of arguments and only their last word must be correctly detected.
Second, they truncate the ACE05 dataset to sentences containing at least one relation both in train and test sets, which leads to an even more favorable setting.

What is worrisome is that both these mistakes are almost invisible in their paper and can only be detected in their code.
The only hint for incorrect evaluation is that they report a score for a setting where they only supervise RE, which is impossible in any standard setting. 
For the dataset, the fact that they do not use the standard preprocessing from \cite{miwa-bansal-2016-end}\footnotemark[1] might be a first clue.

\subsection{Are We Even Using the Same Data?}
Without going this far into data alteration, a first source of ambiguity resides in the use or not of the validation set as additional training data.
While on CoNLL04, because there is no agreement on a dev set, the final model is trained on train+dev by default; the situation is less clear on ACE.
And our following experiments show that this point is already critical w.r.t SOTA claims.

Considering data integrity and keeping the ACE datasets example, even when the majority of works refer to the same preprocessing scripts\footnotemark[1] \footnotetext[1]{\href{https://github.com/tticoin/LSTM-ER}{github.com/tticoin/LSTM-ER}} there is no way to check the integrity of the data without a report of complete dataset statistics.
This is especially true for these datasets whose license prevents sharing of preprocessed versions.

Yet, we have to go back to \cite{roth-yih-2004-linear} to find the original CoNLL04 statistics and \cite{li-ji-2014-incremental} for ACE datasets. 
% And currently available versions of these datasets seem to have different statistics.
To our knowledge, only a few recent works report in-depth datasets statistics \cite{adel-schutze-2017-global,Sanh2019ATasks,Giorgi2019End-to-endModels}.
We report them for CoNLL04 and ACE05 in Table~\ref{table:stats} along with our own.
% To our knowledge, only two more recent works report in-depth datasets statistics: \cite{Sanh2019ATasks} and \cite{Giorgi2019End-to-endModels}.
% \cite{Sanh2019ATasks} reports the exact same statistics as \cite{li-ji-2014-incremental} for ACE05 while in \cite{Giorgi2019End-to-endModels} they differ for both ACE05 and CoNLL04.
% %Despite following the same steps as \cite{Giorgi2019End-to-endModels} (sec.\ref{sec:data}),  we observe differences as shown in Table \ref{table:stats}.

We observe differences in the number of sentences, entity mentions and relations.
Minor differences in the number of annotated mentions likely come from evolutions in datasets versions.
Their impact on performance comparison should be limited, although problematic.   
But we also observe more impactful differences, e.g. with \cite{Giorgi2019End-to-endModels} for both datasets and despite using the same setup and preprocessing.

Such a difference in statistics reminds us that the dataset is an integral part of the evaluation setting.
And in the absence of sufficiently detailed reports, we cannot track when and where they have been changed since their original introduction.

\begin{table}[!h]
\begin{center}
 \resizebox{\columnwidth}{!}{
 \begin{tabular}{@{}ll*{4}{c}@{}}

\toprule
\parbox[t]{2mm}{\multirow{4}{*}{\rotatebox[origin=c]{90}{CoNLL04}}} &  & (R\&Y, 04) & (A\&S, 17) & (G, 19) & Ours\\
\cline{3-6}
& \# sents  &  1,437 &  -       & - & 1,441\\
& \# ents   &  5,336  &   5,302          & 14,193 & 5,349\\
& \# rels  &  2,040 &  2,043             & 2,048 & 2,048\\ 

\midrule
\parbox[t]{2mm}{\multirow{4}{*}{\rotatebox[origin=c]{90}{ACE05}}}&   &   (L\&J, 14) & (S, 19)  &  (G, 19) & Ours \\
\cline{3-6}
& \# sents  &  10,573     &     10, 573       & -         & 14,521 \\
& \# ents   &  38,367      & 34,426 & 38,383    & 38,370 \\
& \# rels   &  7,105       & 7,105 & 6,642     & 7,117 \\

\bottomrule
 \end{tabular}
}
\end{center}
\caption{Global datasets statistics in CoNLL04 and ACE05 as reported by different sources. More detailed statistics are available in Appendix.}
\label{table:stats}
\end{table}

\section{A Small Empirical Study}
%VINCENT \subsection{Motivation}
Given these previous inconsistencies, we can legitimately wonder what is the impact of different evaluation settings on quantitative performance.
However, it is also unrealistic to reimplement and test each and every paper in a same setting to establish a benchmark.
Instead, we propose a small empirical study to quantify the impact of using the Boundaries setting instead of the Strict setting on the two main benchmarks: CoNLL04 and ACE05.
We discard the Relaxed setting because it cannot evaluate true end-to-end RE without strictly taking argument detection into account.
It is also limited to CoNLL04 and we have no example of misuse.

We will consider a limited set of models representative of the main Entity Filtering approach.
And we seize this opportunity to perform two ablations % we think are missing in related work.
that correspond to meaningful recent proposals and are missing in related work.

First, when looking at Table~\ref{table:re_sota}, it is difficult to draw general conclusions beyond the now established improvements due to LM pretraining.
And in the absence of ablation studies on the matter\footnotemark[1], it is impossible to compare models using LM pretraining and anterior works.
For example, in the novel work of \citet{li-etal-2019-entity}, we cannot disentangle the quantitative effects of LM pretraining and the proposed MultiTurn QA. 

Second, to our knowledge, no article compares the recent use of span-level NER instead of classical sequence tagging in end-to-end RE.
%V2
And while Span-level NER does seem necessary to detect overlapping or nested mentions, we can wonder if it is already beneficial on datasets without overlapping entities (like CoNLL04 and ACE05), as suggested by \cite{Dixit2019Span-LevelExtraction}.

\footnotetext[1]{Excepting in \cite{Sanh2019ATasks} which ablates ELMo}

\newcommand{\raisemath}[1]{\mathpalette{\raisemeth{#1}}}
\newcommand{\raisemeth}[3]{\raisebox{#1}{$#2#3$}}
\newcommand{\stddev}{\raisemath{-3pt}}{}

\renewcommand{\arraystretch}{1.4}

\begin{table*}[h]
\centering
\small
\begin{tabularx}{\textwidth}{@{}llr*{2}{Y}r*{2}{Y}r*{2}{Y}r*{2}{Y}r*{2}{Y}r*{2}{Y}r@{}}

\toprule

\multicolumn{3}{c}{\multirow{3}{*}{$\mu$F1}} & \multicolumn{8}{c}{CoNLL04} & & \multicolumn{8}{c}{ACE05}\\
\cline{4-11} \cline{13-20}
 & & & \multicolumn{2}{c}{NER} & & \multicolumn{2}{c}{RE (S)} &  & \multicolumn{2}{c}{RE (B)}&  & \multicolumn{2}{c}{NER} & & \multicolumn{2}{c}{RE (S)} & & \multicolumn{2}{c}{RE (B)}\\

\cline{4-5} \cline{7-8} \cline{10-11} \cline{13-14} \cline{16-17} \cline{19-20}

& & & Dev & Test & & Dev & Test & & Dev & Test & & Dev & Test & & Dev & Test & & Dev & Test\\
\midrule
\parbox[t]{2mm}{\multirow{4}{*}{\rotatebox[origin=c]{90}{BERT}}}
    &\parbox[t]{2mm}{\multirow{2}{*}{\rotatebox[origin=c]{90}{Span}}}
    & train
    & 85.2$_{\stddev{1.9}}$ & 86.5$_{\stddev{1.4}}$ &
    & 69.5$_{\stddev{1.9}}$ & \textbf{67.8}$_{\stddev{.6}}$ &
    & 69.6$_{\stddev{2.0}}$ & \textbf{68.0}$_{\stddev{.5}}$ & 
    
    & 84.6$_{\stddev{.6}}$ & 86.2$_{\stddev{.4}}$ &
    & \textbf{60.1}$_{\stddev{1.0}}$ & \textbf{59.6}$_{\stddev{1.0}}$ &
    &\textbf{63.2}$_{\stddev{.9}}$ &\textbf{62.9}$_{\stddev{1.2}}$ \\
    
    & 
    & +dev 
    & - & 87.5$_{\stddev{.8}}$ &
    & - & \textbf{70.1}$_{\stddev{1.2}}$ & 
    & - & \textbf{70.4}$_{\stddev{1.2}}$ &
    
    & - & 86.5$_{\stddev{.4}}$ &
    & - & \textbf{61.2}$_{\stddev{1.3}}$ &
    & - &\textbf{64.2}$_{\stddev{1.3}}$ \\ 

    & \parbox[t]{2mm}{\multirow{2}{*}{\rotatebox[origin=c]{90}{Seq}}} 
    & train
    & \textbf{86.4}$_{\stddev{1.0}}$ & \textbf{87.4}$_{\stddev{.8}}$ &
    & \textbf{71.0}$_{\stddev{1.8}}$ & \textbf{68.3}$_{\stddev{1.9}}$ &
    & \textbf{71.1}$_{\stddev{1.7}}$ & \textbf{68.5}$_{\stddev{1.8}}$ &
    
    & \textbf{85.7}$_{\stddev{.2}}$ & \textbf{87.0}$_{\stddev{.3}}$ &
    & \textbf{60.1}$_{\stddev{.8}}$ & \textbf{59.7}$_{\stddev{1.1}}$ &
    & 62.6$_{\stddev{1.1}}$ & \textbf{62.9}$_{\stddev{1.2}}$ \\
    
    & 
    & +dev 
    & - & \textbf{88.9}$_{\stddev{0.6}}$ &
    & - & \textbf{70.0}$_{\stddev{1.2}}$ & 
    & - & \textbf{70.2}$_{\stddev{1.2}}$ &
    
    & - & \textbf{87.4}$_{\stddev{.3}}$ &
    & - & \textbf{61.2}$_{\stddev{1.1}}$ &
    & - & \textbf{64.4}$_{\stddev{1.6}}$ \\

\cdashline{1-20}

\parbox[t]{2mm}{\multirow{4}{*}{\rotatebox[origin=c]{90}{BiLSTM}}}
    & \parbox[t]{2mm}{\multirow{2}{*}{\rotatebox[origin=c]{90}{Span}}}
    & train
    & 79.8$_{\stddev{1.6}}$ & 80.3$_{\stddev{1.2}}$ &
    & 61.0$_{\stddev{1.2}}$ & 56.1$_{\stddev{1.4}}$ &
    & 61.2$_{\stddev{1.1}}$ & 56.4$_{\stddev{1.4}}$ &
    
    & 80.0$_{\stddev{.2}}$ & 81.3$_{\stddev{.4}}$ &
    & 46.5$_{\stddev{.8}}$ & 49.4$_{\stddev{1.3}}$ &
    & 49.3$_{\stddev{.9}}$ & 51.9$_{\stddev{1.3}}$ \\
    
    & 
    & +dev 
    & - & 82.7$_{\stddev{1.2}}$ &
    & - & 58.2$_{\stddev{1.5}}$ & 
    & - & 58.5$_{\stddev{1.6}}$ &
    
     & - & 82.2$_{\stddev{.3}}$ &
    & - & 49.3$_{\stddev{.2}}$ &
    & - & 51.9$_{\stddev{.6}}$ \\

    & \parbox[t]{2mm}{\multirow{2}{*}{\rotatebox[origin=c]{90}{Seq}}} 
    & train
    & 80.5$_{\stddev{.7}}$ & 82.0$_{\stddev{.3}}$ &
    & 62.8$_{\stddev{.6}}$ & 60.6$_{\stddev{1.9}}$ &
    & 63.3$_{\stddev{.9}}$ & 60.7$_{\stddev{1.8}}$ &
    
    & 80.8$_{\stddev{.5}}$ & 82.5$_{\stddev{.4}}$ &
    & 47.2$_{\stddev{.5}}$ & 50.3$_{\stddev{1.4}}$ &
    & 49.3$_{\stddev{.5}}$ & 52.8$_{\stddev{1.4}}$ \\
    
    & & +dev 
    & - & 82.6$_{\stddev{.9}}$ &
    & - & 61.6$_{\stddev{1.8}}$ & 
    & - & 61.7$_{\stddev{1.6}}$ & 
    
    & - & 82.8$_{\stddev{.2}}$ &
    & - & 50.1$_{\stddev{1.4}}$ & 
    & - & 52.9$_{\stddev{1.6}}$ \\

\bottomrule

\end{tabularx}

\caption{Double ablation study of BERT and Span-level NER. We report the average of five runs and their standard deviation in subscript. For RE we consider both the Strict and Boundaries settings, RE Strict score is used as the criterion for early stopping.}
\label{table:experiments}

\end{table*}

\subsection{Dataset preprocessing and statistics}
\label{sec:data}
% As highlighted in section 4.4, it is important to clearly state dataset preprocessing and statistics.
We use the standard preprocessing from \cite{miwa-bansal-2016-end} to preprocess ACE05\footnotemark[2].

For CoNLL04, we take the preprocessed dataset and train / dev / test split from \cite{Eberts2020Span-basedPre-training}\footnotemark[3] and check that it corresponds to the standard train / test split from \cite{gupta-etal-2016-table}\footnotemark[4].
We report global dataset statistics in Table \ref{table:stats}.

\footnotetext[2]{\href{https://github.com/tticoin/LSTM-ER}{github.com/tticoin/LSTM-ER}}
\footnotetext[3]{\href{https://github.com/markus-eberts/spert}{github.com/markus-eberts/spert}}  
\footnotetext[4]{\href{https://github.com/pgcool/TF-MTRNN}{github.com/pgcool/TF-MTRNN}}

\subsection{Models}
%V2
We propose to use a model inspired by \cite{Eberts2020Span-basedPre-training} as a baseline for our ablation study since they combine BERT finetuning and Span-level NER.
We then perform two ablations: replacing BERT by a BiLSTM encoder with non-contextual representations and substituting Span-level NER with BILOU sequence tagging.

\paragraph{Encoder : BiLSTM vs BERT}
We use BERT \cite{Devlin-etal-2019-bert} as LM pretraining baseline, expecting that the effects of ELMo \cite{peters-etal-2018-deep} would be similar.
As in related work, we use cased BERT$_{\text{BASE}}$ and finetune its weights.
A word is represented by max-pooling of the last hidden layer representations of all its subwords.

For our non-contextual baseline, we take the previously ubiquitous BiLSTM encoder and choose a 384 hidden size in each direction so that the encoded representation matches BERT's dimension.
We feed this encoder with the concatenation of 300d GloVe 840B word embeddings \cite{pennington-etal-2014-glove} and a reproduction of the charBiLSTM from \cite{lample-etal-2016-neural} (100d char embeddings and hidden size 25 in each direction).

\paragraph{NER Decoder : BILOU vs Span}
In the sequence tagging version, we simply feed the previously encoded word representation $\textbf{h}_i$ into a linear layer with a softmax to predict BILOU tags.
\begin{equation}
    \hat{\textbf{y}}^{seq}_i = \text{softmax}(W^{seq} . \textbf{h}_i + \textbf{b}^{seq})
\end{equation}

For span-level NER, 
% as in \cite{Eberts2020Span-basedPre-training},
we only consider spans up to maximal length 10, which are represented by the max pooling of the representations of their tokens.
An additional span width embedding \textbf{w} of dimension 25 is concatenated to this representation as in \cite{lee-etal-2017-end}.
The only difference with \cite{Eberts2020Span-basedPre-training} is that they also concatenate the representation of the [CLS] token in all span representations to incorporate sentence-level information.
We discard this specificity of BERT-like models.
All these span-level representations are classified using a linear layer followed by a softmax to predict entity types (including None).
We also use negative sampling by randomly selecting 100 negative spans during training.
\begin{align}
    \textbf{h}(s) & = \text{MaxPool}(\textbf{h}_{i}, ...\textbf{h}_{i+l-1})\\
    \textbf{e}(s) & = \lbrack \textbf{h}(s); \textbf{w} (l) \rbrack\\
    % \textbf{e}(s) & = \lbrack \textbf{h}(s); \textbf{\phi }(l) \rbrack \\
    \hat{\textbf{y}}^{span}(s) & = \text{softmax}(W^{span} . \textbf{e}(s) + \textbf{b}^{span})
\end{align}
The NER loss $\mathcal{L}_{NER}$ is the cross-entropy over either BILOU tags or entity classes.

\paragraph{RE Decoder}
For the RE Decoder, we first filter candidate entity pairs i.e. all the ordered pairs of entity mentions detected by the NER decoder.
Then, for every pair, the input of the relation classifier is the concatenation of each span representation \textbf{e}(s$_{i}$) and a context representation \textbf{c}(s$_{1}$, s$_{2}$), the max pooling of all tokens strictly between the two spans\footnotemark[1].
\footnotetext[1]{If there are none, $\textbf{c}(s_{1}, s_{2}) = \bm{0}$}
Once again, this pair representation is fed to a linear classifier but with a sigmoid activation so that multiple relations could be predicted for each pair.
\begin{align}
    \textbf{x}(s_1, s_2) & = \lbrack \textbf{e}(s_1); \textbf{e}(s_2); \textbf{c}(s_{1}, s_{2}) \rbrack \\
    \hat{\textbf{y}}^{rel}(s_1, s_2) & = \sigma (W^{rel} . \textbf{x}(s_1, s_2)  + \textbf{b}^{rel})
\end{align}
$\mathcal{L}_{RE}$ is computed as the binary cross-entropy over relation classes. 
During training, we sample up to 100 random negative pairs of detected or ground truth spans, 
% as in \cite{Eberts2020Span-basedPre-training}.
%%V2
which is different from \cite{Eberts2020Span-basedPre-training} in which negative samples contain only ground truth spans.

\paragraph{Joint Training}
As in most related work, we simply optimize for $\mathcal{L} = \mathcal{L}_{NER} + \mathcal{L}_{RE}$.

\subsection{Experimental Setting}
We implement these models with Pytorch \cite{Paszke2019PyTorch:Library} and Huggingface Transformers \cite{Wolf2019Transformers:Processing}.
For all settings, we fix a dropout rate of 0.1 across the entire network, a 0.1 word dropout for Glove embeddings and a batch size of 8.
We use Adam optimizer \cite{Kingma2015Adam:Optimization} with $\beta_1=0.9$ and $\beta_2=0.999$.
A preliminary grid search on CoNLL04 led us to select a learning rate of  10$^{-5}$ when using BERT and 5.10$^{-4}$ with the BiLSTM\footnotemark[2].

\footnotetext[2]{Search in $\{10^{-6}, 5.10^{-6}, \bm{10^{-5}}, 5.10^{-5}, 10^{-4}\}$ with BERT and $\{10^{-4}, \bm{5.10^{-4}}, 10^{-3}, 5.10^{-3}, 10^{-2}\}$ otherwise.} 

We perform early stopping with patience 5 on the dev set Strict RE $\mu$ F1 score with a minimum of 10 epochs and a maximum of 100.
To compare to related work on CoNLL04, we retrain on train+dev for the optimal number of epochs as determined by early stopping.\footnotemark[3]
% \footnotemark[3].
%V2
\footnotetext[3]{This is not a reproduction of the experimental setting used in \cite{Eberts2020Span-basedPre-training}.}

We report aggregated results from five runs in Table \ref{table:experiments}.

% \footnotetext[3]{Which differs for each random seed.}

\subsection{Comparing Boundaries and Strict Setups}
\label{sec:boundvsstrict}
This humble study first quantifies the impact of using Boundaries instead of Strict evaluation to an overestimation of 2.5 to 3 F1 points on ACE05, which is far from negligible.
% This is far from negligible and should raise awareness on this concern.

But it is also interesting to see that such a mistake has almost no impact on CoNLL04, which highlights an overlooked difference between the two datasets.
A simple explanation is the reduced number of entity types (4 against 7) which reduces the chance to wrongly type an entity.
But we can also notice the difference in the variety of argument types in each relation.
Indeed, in CoNLL04 there is a bijective mapping between a relation type and the ordered types of its arguments; this minimal difference suggests that our models have mostly learned it.
On the contrary % VINCENT, while the NER F1 scores are comparable 
on ACE05, this mapping is much more complex (e.g. the relation PART-WHOLE fits 9 pairs of types\footnotemark[4])\footnotetext[4]{see additional details in Appendix}
which explains the larger difference between metrics, whereas the NER F1 scores are comparable.

\subsection{Comments on the Ablations}
%V2
We must first note that with our full BERT and Span NER baseline, our results do not match those reported by \citet{Eberts2020Span-basedPre-training}.
This can be explained by the slight differences in the models but most likely in the larger ones in tranining procedure and hyperparameters.
Furthermore, we generally observe an important variance over runs, especially for RE.

As expected, the empirical gains mainly come from using BERT, which %enables using simpler decoders for both NER and RE.
allows the use of simpler decoders for both NER and RE.
Indeed, although our non-contextual BILOU model matches \cite{bekoulis-etal-2018-adversarial} on CoNLL04, the results on ACE05 are %under 
overtaken by
models using external syntactic information or more sophisticated decoders with a similar BiLSTM encoder.

% What is more interesting is the comparison between Span-level and sequence tagging NER.
Comparing the Span-level and sequence tagging approaches for NER is also interesting.
%Indeed, a
Although an advantage of Span-level NER is the ability to detect overlapping mentions, its contribution to end-to-end RE on non-overlapping mentions has never been quantified to our knowledge.
Our experiments suggest that it is not beneficial in this case compared to the more classical sequence tagging approach.

\section{How to Prevent Future Mistakes?}

The accumulation of mistakes and invalid comparisons should raise questions to both authors and reviewers of end-to-end RE papers.
How was it possible to make them in the first place and not to detect them in the second place?
How can we reduce their chance to occur in the future?
% We will now expose our humble opinion on the answers to these questions.

\subsection{Lack of Reproducibility}

First, it is no secret that the lack of reproducibility is an issue in science in general and Machine Learning in particular, but we think this is a perfect illustration of its symptoms.
Indeed, in the papers we studied, we only found comparisons to reported scores and rarely an attempt to reimplement previous work by different authors.
This is perfectly understandable given the complexity of such a reproduction, in particular in the multitask learning setting of end-to-end RE and often without (documented) source code.

However, this boils down to comparing results obtained in different settings.
We believe that simply evaluating an implementation of the most similar previous work enables to detect differences in metrics or datasets.
But it also allows to properly assess the source of empirical gains  \cite{Lipton2018TroublingScholarship} which could come from different hyperparameter settings \cite{Melis2018OnModels} or %the proposed 
in-depth changes in the model.

\subsection{Need for More Complete Reports}
Although it is often impossible to exactly reproduce previous results even when the source code is provided, we should at least expect that the evaluation setting is always strictly reproduced.
This requires a complete explicit formulation of the evaluation metrics %explicitly and completely reporting the evaluation metrics
associated with a clear and unambiguous terminology, to which end we advocate for using \cite{bekoulis-etal-2018-adversarial}'s. 
% ; that is why we advocate for using %the clear and unambiguous terminology from
% \cite{bekoulis-etal-2018-adversarial}'s.
Datasets preprocessing and statistics should also be reported
% as completely as possible 
to provide a sanity check.
This should include at least the number of sentences, entity and relation mentions as well as the details of train / test partitions.

\subsection{Towards a Unified Evaluation Setting}

Finally, in order to reduce confusion, we should aim at unifying our evaluation settings.
We propose to always at least report RE scores with the Strict criterion, which considers both the boundaries and types of arguments.
This view matches the NER metrics and truly assess end-to-end RE performance. 
It also happens to be the most used in previous work.

The Boundaries setting proposes a complementary measure of performance more centered on the relation.
The combination of Strict and Boundaries metrics can thus provide additional insights on the models, as discussed in section \ref{sec:boundvsstrict} where we deduce that models can learn the bijective mapping between argument and relation types in CoNLL04.
However, we believe this discussion on their specificities often lacks in articles where both metrics are reported mostly in order to compare to previous works.
Hence we can only encourage to also report a Boundaries score provided sufficient explanation and exploitation of both metrics.

On the contrary, in our opinion, the Relaxed evaluation, which does not account for argument boundaries, cannot evaluate end-to-end RE since it reduces NER to Entity Classification. Furthermore, some papers report the average of NER and RE metrics \cite{adel-schutze-2017-global, Giorgi2019End-to-endModels}, which we believe is also an incorrect metric since the NER performance is already measured in the RE score.

Using a unified setting would also ease cross-dataset analyses and help to better reflect their often overlooked specificities.

\section{Conclusion}

The multiplication of settings in the evaluation of end-to-end Relation Extraction makes the comparison to previous work difficult.
Indeed, in this confusion, numerous articles present unfair comparisons, often overestimating the performance of their proposed model.
Furthermore, this fragmentation of the community complicates the emergence of new proposals.
Our critical literature review epitomizes the need for more rigorous reports of evaluation settings, including detailed datasets statistics.
And we call for a unified end-to-end RE evaluation setting to prevent future mistakes and enable more meaningful cross-domain comparisons.

%V2
Finally, while this article focuses on the necessity to maintain correctness in comparisons and benchmarks, we also believe that further studies are helpful to better understand the behaviors of models.
For example, several works show that lexical overlap in span-based tasks plays a determining role in final performance \cite{moosavi-strube-2017-lexical, Augenstein2017GeneralisationAnalysis, Fu2020RethinkingStudy, Taille2020ContextualizedGeneralization} and others exhibit shallow heuristics in neural Relation Extraction models \cite{Rosenman2020ExposingData, Peng2020LearningExtraction}.

\section*{Acknowledgments}
We thank the anonymous reviewers for their thoughtful and constructive comments.
We thank Markus Eberts for his observations regarding differences with his model and our comparison of NER approaches. 
We thank Giannis Bekoulis, Kalpit Dixit, Pankaj Gupta, Yi Luan, Makoto Miwa, Dat Quoc Nguyen and Victor Sanh for answering our questions on their evaluation settings.

\bibliography{references,anthology}
\bibliographystyle{acl_natbib}

% Appendix

\clearpage
\newpage
\appendix

\renewcommand{\arraystretch}{1.2}

\section{Additional Implementation Details}
We used an Nvidia V100 server with 16BG VRAM for our experiments. 
They can be run with a single Nvidia GTX 1080 with 8GB VRAM with the same hyperparameters as experimented during prototyping.
We report the average number of epochs and time for every configuration in Table \ref{table:app_time}.
We report the number of parameters in our models in Table \ref{table:app_params}.

\begin{table}[h]
\centering
\small
\begin{tabularx}{\columnwidth}{@{}lYRlYR@{}}

\toprule
            & \multicolumn{2}{c}{CoNLL04} & & \multicolumn{2}{c}{ACE05}  \\
            \cline{2-3} \cline{5-6}
Model       & Ep. &  Time & &  Ep. &  Time \\

\midrule

BERT + Span & 52 & 166 & & 25 & 160\\
BERT + BILOU& 16 & 20 & & 22 & 50\\
BiLSTM + Span & 20 & 52 & & 17 & 100\\
BiLSTM + BILOU & 14 & 7 & & 14 & 18\\

\bottomrule

\end{tabularx}

\caption{Average number of epochs before early stopping and corresponding runtime in minutes for a training with early stopping on the dev RE Strict $\mu$ F1 score.}

\label{table:app_time}

\end{table}

\begin{table}[h]
\small
\begin{tabularx}{\columnwidth}{@{}lRR@{}}

\toprule
           
Module      &   CoNLL04 &  ACE05  \\
\midrule

BERT Embedder & 108 M & 108 M\\
GloVe Embedder & 2.6 M & 5.6 M\\
charBiLSTM  & 34 k & 35 k\\
\midrule
BiLSTM Encoder & 2.3 M & 2.3 M\\
\midrule
Span NER & 4 k & 7 k \\
BILOU NER & 13 k & 22 k\\
\midrule
RE Decoder & 12 k & 14 k \\

\bottomrule
BERT + Span & 108 M & 108 M \\
BERT + BILOU & 108 M & 108 M \\
BiLSTM + Span & 5 M & 8 M \\
BiLSTM + BILOU & 5 M & 8 M \\
\bottomrule

\end{tabularx}

\caption{Number of parameters in the different modules of our models.}
\label{table:app_params}
\end{table}

\newpage
\section{Additional Datasets Statistics}
We provide more detailed statistics on the two datasets we used for our experimental study in Tables~\ref{table:app_data_stats_conll04} and \ref{table:app_data_stats_ace05}.
We believe that reporting the number of sentences, entity mentions and relation mentions per training partition is a minimum to enable sanity checks ensuring data integrity.

\begin{table}[h]
\small
\begin{tabularx}{\columnwidth}{@{}ll*{4}{R}@{}}

\toprule
            & Reference & Train & Dev & Test & Total  \\
\midrule
Sentences & (R\&Y, 04)  &   -   & -   &  -   &  1437 \\
            & (G, 16) & 922 & 231 & 288 & 1441 \\
            & Ours &  922 & 231 & 288 & 1441 \\
\midrule
Tokens    & (A\&S, 17) & 23,711 & 6,119 & 7,384 & 37,274\\
          &Ours & 26,525 & 6,993 & 8,336 & 41,854 \\
\midrule
Entities &  (R\&Y, 04)  &   -   & -   &  -   & 5,336 \\
        &  (A\&S, 17) & 3,373 & 858 & 1,071 & 5,302\\
        & Ours & 3,377 & 893 & 1,079 & 5,349 \\
\midrule
Relations &  (R\&Y, 04)  & - & - & - & 2,040  \\
          & (A\&S, 17) & 1,270 & 351 & 422 & 2,043 \\    
          & Ours & 1,283 & 343 & 422 & 2,048\\
\bottomrule

\end{tabularx}

\caption{Detailed statistics of our CoNLL04 dataset, as preprocessed by \citet{Eberts2020Span-basedPre-training} \footnotemark[1].
We compare to previously reported statistics \cite{roth-yih-2004-linear,gupta-etal-2016-table,adel-schutze-2017-global}.
The test sets from \cite{gupta-etal-2016-table}, \cite{adel-schutze-2017-global} and \cite{Eberts2020Span-basedPre-training} are supposedly the same but we observe differences.
Only \cite{Eberts2020Span-basedPre-training} released their complete training partition.}

\label{table:app_data_stats_conll04}

\end{table}

\begin{table}[h]
\centering
\small
\begin{tabularx}{\columnwidth}{@{}ll*{4}{R}@{}}

\toprule
            & Reference & Train & Dev & Test & Total  \\
\midrule
Documents   & (L\&J, 14) & 351 & 80 & 80 & 511\\
            & Ours & 351 & 80 & 80 & 511\\
\midrule
Sentences   & (L\&J, 14)  &  7,273 & 1,765 & 1,535  & 10,573 \\
            
            & Ours & 10,051 & 2,420 & 2,050 & 14,521 \\
\midrule
Tokens    & Ours & 144,783 & 35,548 & 30,595 & 210,926 \\
\midrule
Entities & (L\&J, 14) &  26,470   & 6,421   &  5,476  & 38,367 \\
         & Ours & 26,473 & 6,421 & 5,476 & 38,370 \\
\midrule
Relations & (L\&J, 14)  & 4,779 & 1,179 & 1,147 & 7,105 \\
          & Ours & 4,785 & 1,181 & 1,151 & 7,117\\
\bottomrule

\end{tabularx}

\caption{Detailed statistics of our ACE05 dataset, following \citet{miwa-bansal-2016-end}'s preprocessing scripts\footnotemark[2]. 
We compare to previously reported statistics by \cite{li-ji-2014-incremental}.
The large difference in the number of sentences is likely due to a different sentence tokenizer.
}
\label{table:app_data_stats_ace05}
\end{table}
\footnotetext[1]{\href{https://github.com/markus-eberts/spert}{github.com/markus-eberts/spert}}
\footnotetext[2]{\href{https://github.com/tticoin/LSTM-ER}{github.com/tticoin/LSTM-ER}}

\newpage

\section{Additional Comparison of ACE05 and CoNLL04}
\begin{figure}[h]
    \centering
    \includegraphics[width=0.93\columnwidth]{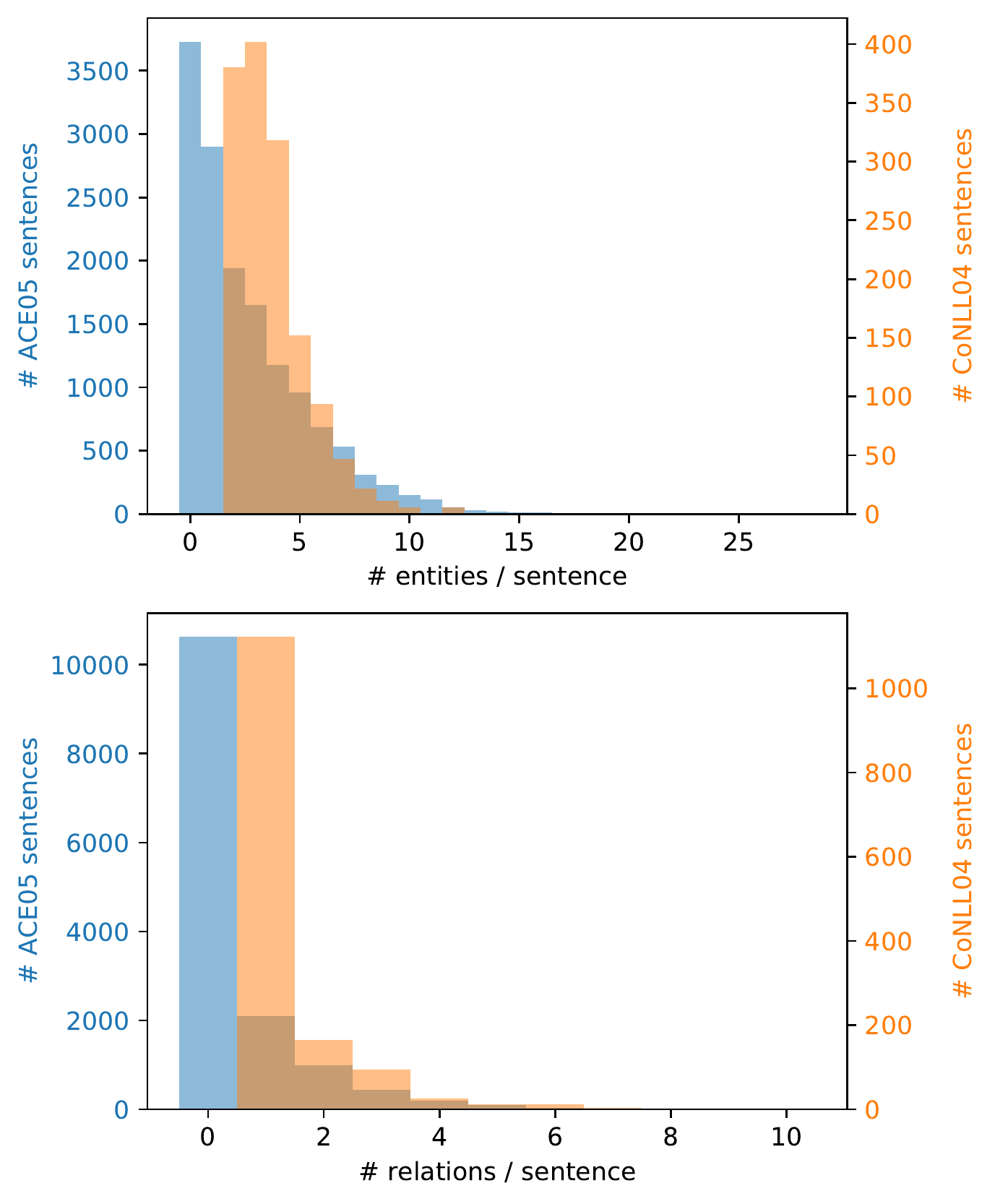}
    \caption{Distribution of the number of entity and relation mentions per sentence in ACE05 and CoNLL04.}
    \label{fig:add_ent_rel}
\end{figure}

\begin{figure}[h]
    \centering
    \includegraphics[width=0.785\columnwidth]{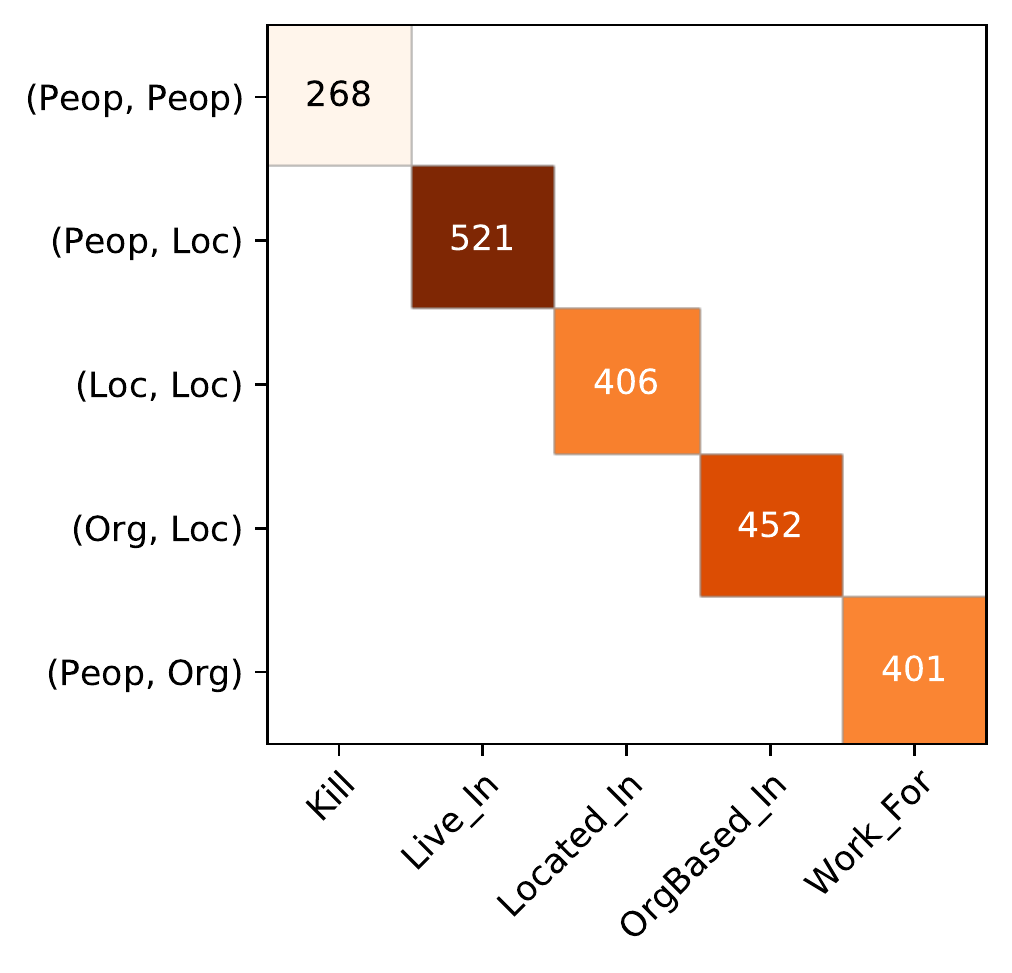}
    \caption{Occurrences of each relation / argument types combination in CoNLL04.}
    \label{fig:add_conll04}
\end{figure}

\begin{figure}[t]
    \centering
    \includegraphics[height=0.95\textheight]{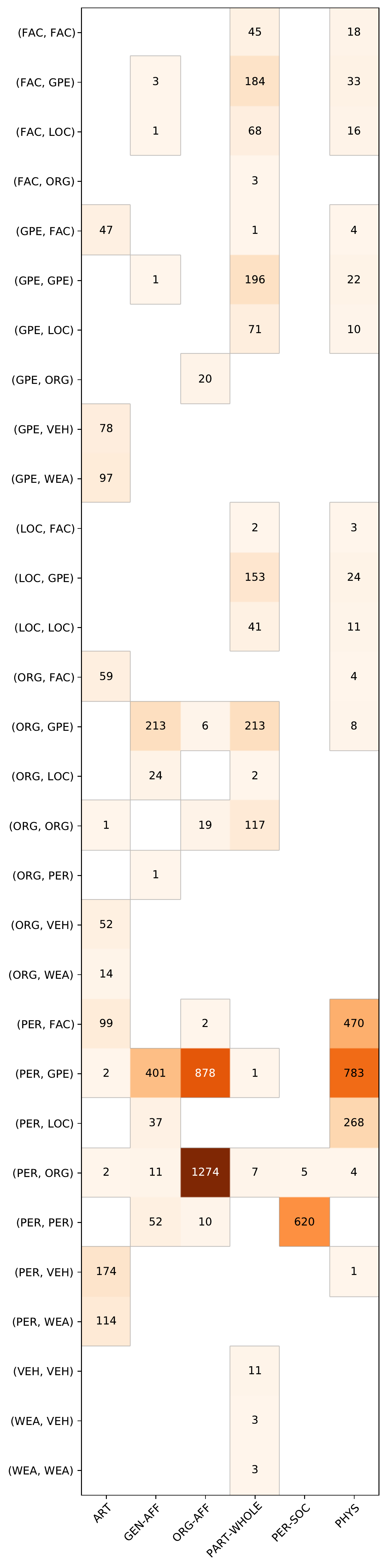}
    \caption{Occurrences of each relation / argument types combination in ACE05.}
    \label{fig:add_ace05}
\end{figure}

ACE05 and CoNLL04 have key differences we propose to visualize with global statistics.
% We propose several additional graphics to highlight key differences between ACE05 and CoNLL04 datasets. 
% This show statistics of the aggregation of all training partitions.
First, in CoNLL04 every sentence contains at least two entity mentions and one relation while the majority of ACE05 contains no entities nor relations as depicted in Fig.~\ref{fig:add_ent_rel}.We can also notice that among sentences containing relations, a higher proportion of ACE05 contain several of them.
Second, the variety of combinations between relation types and argument types makes RE on ACE05 much more difficult than on CoNLL04 (Fig. \ref{fig:add_conll04} and \ref{fig:add_ace05}).

\newpage

\end{document}